\pdfoutput=1

\documentclass[11pt]{article}

\usepackage[preprint]{acl}

\usepackage{bm}

\usepackage{CJKutf8}
\usepackage{makecell}
\usepackage{graphicx}
\usepackage{amsmath}
\usepackage{algorithm, algpseudocode}
\usepackage{hyperref}
\usepackage{multirow}
\usepackage{booktabs}
\algrenewcommand\algorithmicrequire{\textbf{Input:}}
\algrenewcommand\algorithmicensure{\textbf{Output:}}

\usepackage{times}
\usepackage{latexsym}

\usepackage[T1]{fontenc}

\usepackage[utf8]{inputenc}

\usepackage{microtype}

\usepackage{inconsolata}

\usepackage{graphicx}

\title{Semantic-Augmented Latent Topic Modeling with LLM-in-the-Loop}

\author{Mengze Hong$^{1}$\hspace{0.8ex} Chen Jason Zhang$^{1}$ \hspace{0.8ex} Di Jiang$^{2}$\thanks{Corresponding author: \href{mailto:dijiang@webank.com}{dijiang@webank.com}}\\
$^{1}$Hong Kong Polytechnic University \quad $^{2}$AI Group, WeBank Co., Ltd\\
}

\begin{document}
\maketitle

\begin{abstract}

Latent Dirichlet Allocation (LDA) is a prominent generative probabilistic model used for uncovering abstract topics within document collections. In this paper, we explore the effectiveness of augmenting topic models with Large Language Models (LLMs) through integration into two key phases: Initialization and Post-Correction. Since the LDA is highly dependent on the quality of its initialization, we conduct extensive experiments on the LLM-guided topic clustering for initializing the Gibbs sampling algorithm. Interestingly, the experimental results reveal that while the proposed initialization strategy improves the early iterations of LDA, it has no effect on the convergence and yields the worst performance compared to the baselines. The LLM-enabled post-correction, on the other hand, achieved a promising improvement of 5.86\% in the coherence evaluation. These results highlight the practical benefits of the LLM-in-the-loop approach and challenge the belief that LLMs are always the superior text mining alternative.


\end{abstract}

\section{Introduction}

Topic modeling uncovers latent themes in text corpora, offering undeniable practical values in driving applications such as news recommendation, content analysis, and information retrieval \cite{blei2003latent, li2021heterogeneous, jiang2023probabilistic}. Latent Dirichlet Allocation (LDA) is a widely used generative probabilistic model for discovering abstract topics within a collection of documents, representing documents as mixtures of topics. While it has shown great effectiveness in capturing the underlying relationship between word-topic and topic-document, it faces two significant limitations: (1) \textit{semantically uninformed initialization}, which slows convergence and yields incoherent early topics \cite{jiang2021industrial}, and (2) \textit{poor topic coherence}, where topics often include unrelated terms due to the complex nature of document composition \cite{ABDELRAZEK2023102131}.

The advent of Large Language Models (LLMs) has not only transformed conversational systems \cite{hong2025dialoguelanguagemodellargescale, hong2025expandingchatbotknowledgecustomer} but also opened up new opportunities for tackling a wide range of NLP tasks \cite{viswanathan2023large}. These models provide human-level semantic understanding and contextual awareness, enabling more advanced and effective LLM-native solutions \cite{jiang2021familia, 10.5555/3666122.3666809, song2024communication}. However, directly applying LLMs to topic modeling is hindered by input length constraints, vulnerability to hallucinations, and a lack of deterministic, transparent modeling processes, which obscure topical relationships and reduce interpretability \cite{mu-etal-2024-large}. To overcome these inherent challenges of LLMs, the LLM-in-the-loop methodologies \cite{llm-in-the-loop} demonstrated increasing prominence of integrating LLM utilities to assist the modeling process \cite{stammbach-etal-2023-revisiting, bianchi-etal-2021-pre}. Existing research has demonstrated significant improvements in semantically-rich tasks, such as intent clustering \cite{hongintent2025, viswanathan2023large} and speech recognition (ASR) \cite{asano-etal-2025-contextual, hong2025technicalreportpracticalguide}, yet there are limited efforts in augmenting topic modeling.

In this paper, we investigate the effectiveness of integrating few-shot LLMs to provide semantic guidance across two key stages of topic modeling: Initialization and Post-Correction. Through comprehensive experiments, we highlight the significance of LLM-guided initialization for the Gibbs sampling algorithm used in LDA, reducing the burn-in time and significantly improving the convergence efficiency compared to conventional approaches. The main findings of this paper include:

\begin{enumerate}
    \item The effectiveness of LLMs in replacing topic models has not fully met general expectations due to limitations in instruction following and solution space alignment. More broadly, replacing generative probabilistic models entirely with generative language models presents significant challenges and may not be a universally feasible approach.
    \item Leveraging LLM-guided topical clusters for initializing Gibbs sampling provides some benefits during the initial iterations, but does not significantly impact the model's convergence speed. This finding contrasts with preliminary studies suggesting that initialization would have a more pronounced effect on model efficiency.
    \item Post-correction of topic models using few-shot LLMs proves effective, resulting in a 5.86\% improvement in topic coherence and enhanced interpretability. This highlights the practical benefits of incorporating task-centric LLM utility into the modeling process.
\end{enumerate}

These insights, particularly the negative results, challenge the widespread belief that LLMs are universally superior for all NLP tasks. The use of LLMs in an ``in-the-loop'' approach has demonstrated enhanced performance by effectively combining the strengths of both LLMs and task-specific models like LDA,  showcasing its potential to overcome the limitations of relying solely on LLMs.


\section{Preliminaries}
\subsection{Topic Modeling}
Topic modeling has been well-established as an unsupervised text mining technique for identifying significant topics within a document corpus. This approach involves analyzing patterns in word occurrences to discover themes or topics \cite{blei2003latent, song2021classification, grootendorst2022bertopic, jiang2012g} and outputting a set of topics, each represented by a list of salient words. Topics can be interpreted manually or automatically, often requiring descriptive labels to encapsulate their semantic content \cite{lau2010best}. Additionally, data preprocessing such as stemming and lemmatization can significantly impact the performance of topic modeling \cite{chuang2015topiccheck, jiang2013mining, schofield2016comparing} and its downstream applications \cite{jiang2019federated, vosecky2014integrating}.

\subsection{LDA and Gibbs Sampling}
Latent Dirichlet Allocation (LDA) models documents as mixtures of topics, with topics as distributions over words. The objective is to estimate the document-topic distribution \(\theta_d \in \Delta^{T-1}\), where \(T\) denotes the number of topics, and the topic-word distribution \(\phi_t \in \Delta^{|\mathcal{V}|-1}\), where \(\mathcal{V}\) represents the vocabulary, which reduces the dimensionality of document representation.

Gibbs sampling is a Markov Chain Monte Carlo (MCMC) technique used to approximate the posterior \(P(\mathbf{z} \mid \mathbf{w}, \alpha, \beta)\) by iteratively sampling topic assignments \(z_i\) from the conditional distribution:
\vspace{-0.5em}
\begin{equation*}
\begin{aligned}
&P(z_i = t \mid w_i, \mathbf{z}_{-i}, \mathbf{w}_{-i}, \alpha, \beta) \propto \\&\frac{n_{wt,-i} + \beta}{\sum_{w'} n_{w't,-i} + |\mathcal{V}|\beta} \cdot \frac{n_{td,-i} + \alpha}{\sum_{t'} n_{t'd,-i} + T\alpha},
\end{aligned}
\end{equation*}

\noindent where \(n_{wt,-i}\) is the count of word \(w\) assigned to topic \(t\) excluding the \(i\)-th word, \(n_{td,-i}\) is the count of topic \(t\) in document \(d\) excluding the \(i\)-th word, and \(\alpha, \beta\) are Dirichlet hyperparameters controlling topic sparsity \cite{jiang2016query}.

\subsection{Gibbs Sampling Initialization}
Gibbs sampling for LDA begins by initializing topic assignments \(z_i\) for each word \(w_i\) in the corpus. For a corpus with \(N\) documents and \(M_d\) words per document \(d\), each word \(w_i \in \mathcal{V}\) is assigned a topic \(z_i \in \{1, \dots, T\}\), producing initial counts \(n_{wt}\) and \(n_{td}\). The initialization strategy influences convergence rates of the Markov chain. Common approaches include random assignment, where \(z_i \sim \text{Multinomial}(1/T)\), coin flip assignment for binary topic cases, and adaptive initialization leveraging document clustering or word embeddings \cite{resnik2010gibbs, vosecky2013dynamic, ROBERTS1994207}. The iterative updates ensure convergence to the stationary distribution, approximating the posterior \(P(\mathbf{z}, \theta, \phi \mid \mathbf{w}, \alpha, \beta)\), as guaranteed by the ergodicity of the Markov chain under mild conditions.

\subsection{Topic Distillation of LLM Semantics}

Intuitively, given the robust natural language understanding and generation capabilities of LLMs, one might consider directly substituting traditional topic models with LLMs. In this work, we explore leveraging LLMs to distill corpus semantics into predefined topics. By inputting the corpus vocabulary into an LLM and employing a few-shot prompt, we aim to cluster terms into a specified number of topics \( T \) \cite{mu-etal-2024-large}, capitalizing on LLMs' advanced semantic comprehension \cite{etal-2024-evaluating}.

However, this approach presents significant challenges due to limitations of LLMs in coherently organizing large sets of topics \cite{doi-etal-2024-topic}. Initial experiments reveal that LLM hallucinations and the inconsistent instruction following can substantially hinder performance. For simpler tasks, such as clustering 100 words into 10 topics, LLMs perform comparably to LDA, producing results that align closely with human interpretations. However, as the task complexity increases, such as clustering 1000 words, LLMs exhibit repetitive word assignments and introduce random noise, failing to align with the desired solution space. This emphasizes the limitations of LLM-native solutions and provides strong motivation for developing LLM-in-the-loop approaches to overcome these challenges.

\section{LLM-in-the-loop Topic Modeling}

In this paper, we propose an LLM-in-the-loop topic modeling framework emphasizing the incorporation of LLM utilities into the initialization and post-correction stages.

\subsection{Gibbs Sampling Initialization}

In Gibbs sampling, random topic assignments can be refined using prior knowledge from text clustering to guide the initialization of topic-word distributions. One approach is to leverage text clustering to obtain this prior knowledge. However, traditional methods typically optimize embedding-distance between clusters, providing limited semantic insights \cite{wu-etal-2014-improve}. To better align with human interpretation, LLMs can be integrated to enhance the unsupervised process of preliminary topic clustering, enabling more context-aware groupings.

By employing few-shot LLMs as semantic coherence evaluators, we assess clustering results by computing a coherence score \( C_v = \frac{1}{T} \sum_{t=1}^T \sum_{w_i, w_j \in t} \text{sim}_{\text{LLM}}(w_i, w_j) \), where \(\text{sim}_{\text{LLM}}\) measures semantic similarity between words \(w_i, w_j\) in cluster \(t\). Clusters with \( C_v \) above a threshold are retained, serving as initial topic assignments for Gibbs sampling in LDA. This approach potentially reduces the number of iterations required for convergence to the posterior \( P(\mathbf{z}, \theta, \phi \mid \mathbf{w}, \alpha, \beta) \), while enhancing topic quality, making topics more coherent and reflective of nuanced language patterns identified by the LLM.

\subsection{Post-Correction of Topic Models}

After LDA generates the initial topic assignment, LLMs can be employed to further optimize and refine the topics. Besides the direct topic interpretation, where LLMs are prompted to provide coherent and meaningful topic descriptions, this paper considers word-level post-correction to improve the coherence of the generated topics. This involves filtering out semantically unrelated words and ensuring that the topics are distinct and comprehensive. With this post-correction, the final set of topics is expected to be more accurate and semantically correlated, effectively capturing the underlying themes and improving the overall interpretability of the topic model.



\section{Experiments}
\subsection{Experiment Setup}

We evaluate topic models using a dataset of 10,000 Chinese news articles, integrating the Qwen2-7B-Instruct model at various stages of topic modeling. For comparison, we use the K-Means algorithm for initialization as the baseline. The LDA hyperparameter `eta,' controlling the topic-word distribution prior, is set to `None' for even distribution and `0.1' for sparser distribution. 

\subsection{Metrics}
Two metrics are used to compare the performance of different topic models:

\vspace{1em}
\noindent\textbf{Perplexity} evaluates how well the inferred topics represent the underlying structure of the corpus. A lower perplexity value indicates better predictive performance, and the model's convergence behavior can be tracked by examining the perplexity at each iteration.
\[
PP(p) = 2^{-\sum_x p(x) \log_2 p(x)} = \prod_x p(x)^{-p(x)}
\]

\noindent\textbf{Topic coherence} measures the degree of semantic similarity between the top words in a topic, providing insight into how interpretable and meaningful the topics are to humans \cite{jiang2016cross}. The Normalized Pointwise Mutual Information (NPMI) is used to evaluate the coherence of topics by measuring the association between words.
\[
\text{PMI}(w_i, w_j) = \log \frac{P(w_i, w_j)}{P(w_i) P(w_j)}
\]
\[
\text{NPMI}(w_i, w_j) = \frac{\text{PMI}(w_i, w_j)}{-\log P(w_i, w_j)}
\]

\begin{CJK}{UTF8}{gkai} 
\begin{table*}[t]
\centering
\resizebox{\textwidth}{!}{
\begin{tabular}{|l|l|l|}
\hline
 \thead{\textbf{Evaluation}} & \thead{\textbf{Words}} & \thead{\textbf{English Translation}}  \\ \hline
 \thead{Original} & \thead{宝宝, 妈妈, \textbf{星座}, 玩具, \textbf{宇宙}, \\母乳, \textbf{科学}, 宝贝, 喜欢, 母婴, \\\textbf{发现}, 狮子, 世界, 奶粉, 处女座, \\时间, 双胞胎, 洗澡, \textbf{量子}, 橘子} & \thead{Baby, Mom, \textbf{Constellation}, Toy, \textbf{Universe}, \\Breast milk, \textbf{Science}, Baby, Like, Mother and baby, \\\textbf{Discover}, Lion, World, Milk powder, Virgo, \\Time, Twins, Bath, \textbf{Quantum}, Orange} \\ \hline
 \thead{Filtered} & \thead{宝宝, 妈妈, 玩具, 母乳, 宝贝, \\喜欢, 母婴, 狮子, 世界, 奶粉, \\处女座, 时间, 双胞胎, 洗澡, 橘子} & \thead{Baby, Mom, Toy, Breast milk, Baby, \\Like, Mother and baby, Lion, World, Milk powder, \\Virgo, Time, Twins, Bath, Orange} \\ \hline
\end{tabular}%
}
\caption{Example of LLM-Guided Post-Correction: words in \textbf{bold} are identified by LLM as irrelevant}
\label{tab:LLM_Post_Correction}
\end{table*}
\end{CJK}

\subsection{Results and Discussions}

\begin{table}[!t]
\begin{tabular}{cccc}
\toprule
\textbf{Method} & \textbf{eta} & \textbf{pass = 0} & \textbf{pass = 20} \\ \midrule
\multirow{2}{*}{Random}  & None & 8158.6628 & 309.3647 \\ 
 & 0.1 & 5170.6256 & \textbf{285.0380} \\ \midrule
\multirow{2}{*}{Cluster}  & None & 7911.5573 & 323.1183 \\ 
& 0.1 & 5003.9228 & 301.7249 \\ \midrule
\multirow{2}{*}{LLM (\textbf{ours})} & None & 7853.6795 & 328.9051 \\ 
& 0.1 & \textbf{4967.7951} & 306.4441 \\ \bottomrule
\end{tabular}

\caption{Perplexity evaluation of different approaches}
\label{trainset_perplexity}
\end{table}

\begin{table}[!t]
\centering
\begin{tabular}{cccc}
\toprule
\textbf{Method} & \textbf{eta} & \textbf{pass = 0} & \textbf{pass = 20} \\ \midrule
\multirow{2}{*}{Random} & None & \textbf{-17.6309} & -3.4383 \\ 
 & 0.1 & -17.8018 & \textbf{-3.2723} \\ \midrule
\multirow{2}{*}{Cluster}  & None & -17.7566 & -3.7144 \\ 
 & 0.1 & -17.7925 & -5.2016 \\ \midrule
\multirow{2}{*}{LLM (\textbf{ours})} & None & -17.8434 & -4.3831 \\ 
 & 0.1 & -17.8463 & -5.6805 \\ \bottomrule
\end{tabular}
\caption{Coherence evaluation of different approaches}
\label{trainset_coherence}
\end{table}

\subsubsection{LLM-Guided Initialization}

The evaluation of perplexity is presented in Table \ref{trainset_perplexity}. While the LLM-guided initialization enhances the initial performance of the topic model, it has no effect on the final outcomes and results in the worst performance among the three methods. With closer examination, it is observed that the baseline methods surpass the LLM initialization at the 4th iteration and achieve an average descent rate that is 3\% higher than the LLM approach. 

The coherence evaluation result in Table \ref{trainset_coherence} assesses the quality of the discovered topics and further highlights the ineffectiveness of LLMs in initializing Gibbs sampling. This can be attributed to two main reasons: 1) The initialization with clusters introduced noise and external bias to the topic modeling process \cite{10.1007/3-540-48412-4_18}, and 2) The LLM evaluation may not align well with the inherent structure of the dataset, resulting in sub-optimal topic coherence scores.

\subsubsection{LLM-Guided Post-Correction}

In the post-processing phase, a few-shot LLM is employed to refine the topic collection by eliminating semantically unrelated terms. This method resulted in a 5.86\% improvement in topic coherence, demonstrating the practical advantages of using LLMs to enhance the accuracy and relevance of topic modeling outcomes. Table \ref{tab:LLM_Post_Correction} showcases a case study of LLM-enabled post-correction, where certain words in the original output appeared semantically disconnected from others. The LLM effectively identifies and filters out these unrelated terms, ensuring that the final set of topics remains semantically coherent.

\section{Conclusion}

This study investigates the integration of few-shot LLMs to augment topic models through an LLM-in-the-loop approach. The experimental results reveal the limitations of LLMs, emphasizing the need for careful evaluation when considering generative language models as substitutes for traditional probabilistic models. Despite these limitations, LLMs proved effective in task-centric post-processing, enhancing topic coherence by refining and filtering topic word collections. This demonstrates their strength in improving the interpretability of topic models, aligning with previous research that positions LLM-in-the-loop machine learning as a promising direction. Future research is encouraged to optimize such LLM-in-the-loop integration frameworks and explore their application to a broader range of text mining tasks.



\bibliography{reference}

\appendix








\begin{CJK}{UTF8}{gkai} 
\begin{table*}[!t]
\centering
\resizebox{\textwidth}{!}{
\begin{tabular}{|l|l|l|}
\hline
 \thead{\textbf{Evaluation}} & \thead{\textbf{Words}} & \thead{\textbf{English Translation}}  \\ \hline
 \thead{Good} & \thead{万丈, 两百年, 中百, 十万, 十万元,\\ 十万块, 十百万, 十亿, 十亿美元, 十余, \\十余万, 十余位, 十余名, 十余家, \\十余年, 十余载,  十元, 十块, \\十块钱, 十百} & \thead{Ten thousand feet, Two hundred years, Hundred, Hundred thousand, \\ Hundred thousand yuan, Hundred thousand pieces, Ten million, \\ Billion, Billion dollars, More than ten, More than a hundred thousand, \\ More than ten places, More than ten people, More than ten companies, \\ More than ten years, More than ten years, Ten yuan, \\Ten pieces, Ten dollars, Ten hundred} \\\hline
 \thead{Bad} & \thead{万丈深渊, 两层, 两层楼, 云层, \\刷层, 加厚, 加深} & \thead{Abyss, Two layers, Two floors, Cloud layer, \\Brush layer, Thicken, Deepen} \\\hline
\end{tabular}%
}
\caption{Example of LLM-supervised clustering}
\label{tab:LLM_supervised_clustering}
\end{table*}
\end{CJK}

\section{Prompts}
\label{appendix:prompts}

\begin{itemize}
    \item \textbf{LLM for Topic Inference} \\ \textit{Generate \{count\} set of topics from the given corpus vocabulary. The criteria are: 1) Cluster the terms into \{count\} topics; 2) Ensure each topic is semantically coherent; 3) Avoid repetition of words across topics and minimize random noise. Examples: \{examples\}. Topics: \{topics\}}
    \item \textbf{Coherence Evaluation} \\ \textit{Evaluate whether the following set of words forms a semantically coherent cluster. The criteria are: 1) All words should relate to the same overarching theme or concept; 2) Their meanings and contexts should align closely; 3) There should be no outliers. If the cluster meets these criteria, respond with "Yes." If not, respond with "No". Examples: \{examples\}. cluster to Evaluate: \{cluster\}}
    \item \textbf{Post-correction} \\ \textit{Assess whether the following Chinese words meet the specified criteria. The criteria are: 1) All words should belong to the same category or concept; 2) The meanings, contexts, and usage should be very similar; 3) No exceptions should exist. If the words meet the criteria, respond with "Yes." If the words do not meet the criteria, respond with "No" and list the words that do not fit using Python list syntax. Examples: \{examples\}. Words to Evaluate: \{words\}}
\end{itemize}



\end{document}